\documentclass[11pt,a4paper]{article}
\usepackage[hyperref]{acl2020}
\usepackage{amsmath,amsfonts,bm}

\def\eqref#1{equation~\ref{#1}}
\def\1{\bm{1}}

\def\vh{{\bm{h}}}

\DeclareMathAlphabet{\mathsfit}{\encodingdefault}{\sfdefault}{m}{sl}
\SetMathAlphabet{\mathsfit}{bold}{\encodingdefault}{\sfdefault}{bx}{n}

\usepackage[vlined,ruled,linesnumbered]{algorithm2e}
\usepackage{times}
\usepackage{booktabs}
\usepackage{hyperref}
\usepackage{url}
\usepackage{graphicx}
\usepackage{amsmath}
\usepackage{bm}
\usepackage{multirow}
\usepackage{amssymb}
\usepackage{arydshln}
\usepackage{pgfplots}
\usepackage{subfigure}

\definecolor{orange1}{RGB}{255, 145, 78}
\definecolor{blue1}{RGB}{88, 188, 255}
\usepackage{microtype}

\aclfinalcopy % Uncomment this line for the final submission

\title{Accurate Word Representations with Universal Visual Guidance}

\author{Zhuosheng Zhang\textsuperscript{1,2,3}, Haojie Yu\textsuperscript{1,2,3}, Hai Zhao\textsuperscript{1,2,3,\thanks{\ \ Corresponding author.  This paper was partially supported by National Key Research and Development Program of China (No. 2017YFB0304100), Key Projects of National Natural Science Foundation of China (U1836222 and 61733011).}}, Rui Wang\textsuperscript{1,2,3}, Masao Utiyama\textsuperscript{4}, \\
\textsuperscript{1} Department of Computer Science and Engineering, Shanghai Jiao Tong University\\
\textsuperscript{2} Key Laboratory of Shanghai Education Commission for Intelligent Interaction\\
and Cognitive Engineering, Shanghai Jiao Tong University, Shanghai, China\\
\textsuperscript{3}MoE Key Lab of Artificial Intelligence, AI Institute, Shanghai Jiao Tong University, Shanghai, China\\
\textsuperscript{4} National Institute of Information and Communications Technology (NICT)\\
\texttt{\{zhangzs,hudiefeiafei\}@sjtu.edu.cn,zhaohai@cs.sjtu.edu.cn}\\
\texttt{wangrui.nlp@gmail.com, mutiyama@nict.go.jp}\\
}

\date{}

\begin{document}
\maketitle
\begin{abstract}
Word representation is a fundamental component in neural language understanding models. Recently, pre-trained language models (PrLMs) offer a new performant method of contextualized word representations by leveraging the sequence-level context for modeling. Although the PrLMs generally give more accurate contextualized word representations than non-contextualized models do, they are still subject to a sequence of text contexts without diverse hints for word representation from multimodality. This paper thus proposes a visual representation method to explicitly enhance conventional word embedding with multiple-aspect senses from visual guidance. In detail, we build a small-scale word-image dictionary from a multimodal seed dataset where each word corresponds to diverse related images. The texts and paired images are encoded in parallel, followed by an attention layer to integrate the multimodal representations. We show that the method substantially improves the accuracy of disambiguation. Experiments on 12 natural language understanding and machine translation tasks further verify the effectiveness and the generalization capability of the proposed approach.

\end{abstract}

\section{Introduction}

\begin{figure*}[htb]
	\centering
	\includegraphics[width=0.95\textwidth]{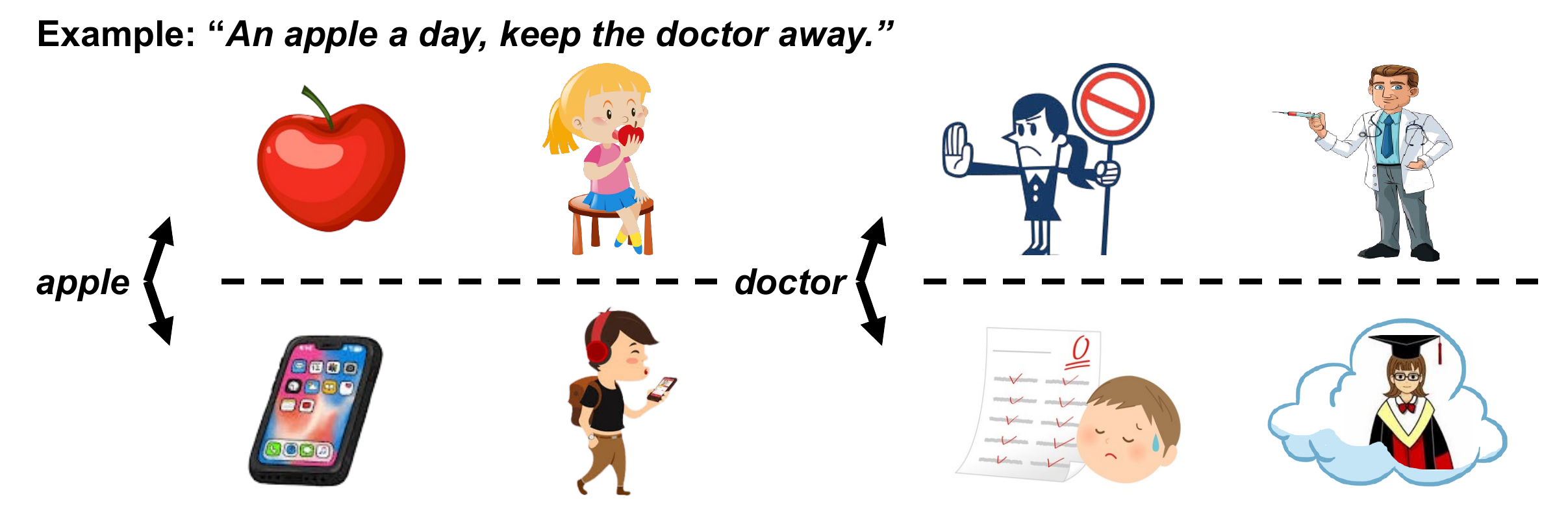}
	\caption{\label{fig:intro_exp} Example of ambiguity. This sentence ambiguity which comes from polysemy ambiguities cannot be solved only by text context.}
\end{figure*}

Learning word representations has been an active research field that has gained renewed popularities for decades \citep{mikolov2013distributed,radford2018improving}. Word representations have been evolved from standard distributed representations \citep{brown1992class,mikolov2013distributed,pennington2014glove} to contextualized language representations from deep pre-trained representation models (PrLMs) \citep{Peters2018ELMO,radford2018improving,devlin2018bert}. The former static embeddings are commonly derived from distributed representations through capturing the local co-occurrence of words from large-scale unlabeled texts. In contrast, the later contextualized representations are mainly obtained by PrLMs. However, the contexts during language modeling are subject to the individual input text sequence, without diverse hints for word representation from multimodality. Polysemy is a common phenomenon in human language. As shown in Figure \ref{fig:intro_exp}, word polysemy may result in completely different sentence meanings.\footnote{There exists sentence-level syntactic ambiguity in languages as an example, \textit{the girl saw the man on the hill with a telescope}, shown in Appendix \ref{appendix-amb}, which is not the focus in this work. We left it in future research.} The difficulty lies in what kind of meanings the word expresses in each context, and it may be insufficient to obtain the exact representations from the sentence context alone. Therefore, in addition to the contextual word representation, we turn to model multiple-aspect representations for each word to refine more accurate sentence representations. 

Recently, the study of multimodal representations has aroused great interest, which shows potential benefits to the language representations through a more comprehensive perception of the real world \citep{zhang2018image}. The major theme of multimodality is to use different modality sources such as texts and images to get a more distinguished representation, which would be beneficial than learning representations from only one modality \citep{lazaridou2015combining}. Multimodal information provides more discriminative inputs such as image features than learning representations from only one modality, which is potential to help deal with the ambiguity of words in text-only modality. Most of previous works focused on jointly modeling images and texts \citep{su2019vl,lu2019vilbert,tan2019lxmert,li2019unicoder,zhou2019unified,sun2019videobert,kiela2014learning,silberer2014learning,zablocki2018learning,wu2019glyce}.
However, these studies rely on text-image annotations as the paired input. They thus are retrained only in vision-language and multimodal tasks, such as image captioning and visual question answering. For natural language processing (NLP) tasks, most text is not distributed with visual information; therefore, it is essential to break the constraint of the annotation prerequisite when applying visual information to a wide range of mono-modal (e.g., text-only) tasks. To solve the bottlenecks of the high cost of manual image annotation, \citet{zhang2020neural} proposed to build a lookup table from a multimodal dataset and then used the search-based method to retrieve images for each input sentence. However, the lookup table is constructed at sentence-level from \textit{Multi30K} \citep{elliott2016Multi30K}, which would lead to the relatively limited coverage of the related images for each sentence, and suffers from noise as well. Different from the previous work that incorporates the visual modality as sentence-level guidance, we propose to leverage word-level multimodal assistance inspired by our observation in Figure \ref{fig:intro_exp}, which offers more accurate means for disambiguation as word-level clues may conveniently help solve sentence-level ambiguity.\footnote{Instead of a linguistic term, ``disambiguation" in this paper means a broad kind of practice for more distinguishable representation from diverse input sources.} Meanwhile, multimodality derived from word representations can also be more helpful to provide as many images as possible to alleviate the scarcity of images retrieved for sentences. Furthermore, word-level operation enables the resultant model to benefit broad NLP tasks more than machine translation as previous work focuses on.

In this paper, we propose a visual representation method to explicitly enhance each word with multiple-aspect senses by the visual modality.  In detail, based on the universal visual representation method \citep{zhang2020neural}, we build a word-image dictionary from the small-scale multimodal dataset Multi30K where one word corresponds to diverse images,\footnote{We describe our method by regarding the processing unit as word though this method can also be applied to a subword-based sentence for which the subword is considered to be the processing unit.} which can further be used for text-only tasks. During training and decoding processes, multiple and diverse images corresponding to the input word, which represent the multiple-aspect senses of the word, will be retrieved from the dictionary and then be encoded as image representations by a pre-trained ResNet \citep{he2016deep}. The texts and paired images are encoded in parallel, followed by an attention layer to fuse those representations. Then, the fused representations are passed to downstream task-specific layers for prediction.
% In particular, we integrate the proposed approach into text-only neural machine translation (NMT) model Transformer \citep{NIPS2017_7181} and evaluate its effectiveness on several translation datasets, including Multi30K, etc. 
Experiments on various natural language understanding and machine translation tasks verified the effectiveness and the generalization capability of the proposed approach.

% {\color{red}【注释的main contribution写的挺好的，这篇文章和已有工作的区别还是说清楚比较好。】}
% In summary, our contributions are primarily three-fold: \\
% 1. We hypothesize that each word naturally has multiple-aspect senses which are difficult to be obtained from the sentence context alone, and thus propose our visual-aware solution accordingly.\\
% 2. We present a simple yet effective visual representation method that explicitly give each word its nature multiple-aspect senses.\\
% 3. Experiments on a various natural language understanding and generation tasks verified the effectiveness and the generality of the proposed approach.

\section{Background}
\subsection{Language Representation}

Training machines to comprehend human language requires comprehensive and accurate modeling of natural language texts, in which the fundamental component is language representation \cite{zhang2020mrc}. Empirically, distributed representations \citep{brown1992class,mikolov2013distributed,pennington2014glove} have been widely applied in diverse NLP tasks \cite{zhang2018dua,li2018unified}, which benefit from the ability to capture the local co-occurrence of words from the large-scale unlabeled text. 
% \textcolor{red}{However, these approaches blindly create word vectors without taking into account the shared dependencies and context in which each word entails.} 
Recently dominant pre-trained contextualized language models such as ELMo \citep{Peters2018ELMO}, GPT \citep{radford2018improving}, and BERT \citep{devlin2018bert} focus on learning context-dependent representations by taking into account the context for each occurrence of a given word \citep{Peters2018ELMO}
% \textcolor{red}{fill the gap by capturing the semantic, contextual, and syntactic meaning of each word in the corpus vocabulary based on the usage of these words in sentences.}
The major technical improvement over the traditional embeddings of these newly proposed language models is that they focus on extracting context-sensitive features, whose contextualized embedding for each word will be different according to the sentence.
When integrating these contextual word embeddings with existing task-specific architectures, ELMo helps boost several major NLP benchmarks \citep{Peters2018ELMO}, including question answering on SQuAD \cite{Rajpurkar2016SQuAD}, sentiment analysis \citep{socher2013recursive}, and named entity recognition \citep{sang2003introduction}. The latest evaluation shows that BERT \citep{devlin2018bert} especially shows effectiveness in language understanding tasks on GLUE, MultiNLI, and SQuAD \citep{zhang2020semantics,zhang2020sg,zhou2020limit,zhang2020dcmn+}. Nevertheless, the word embeddings learned through these approaches are completely based on the textual corpus. In practice, the meaning of the word itself would involve multiple senses and naturally leads to ambiguity. In this work, we focus on more accurate word representation with effective disambiguation for pre-trained contextualized language models.

\begin{algorithm*}
            \caption{Word-image Dictionary Conversion Algorithm}\label{alg}
        	\KwIn {Input sentence set $S=\{X_1, X_2, \dots\, X_k\}$ and paired image set $E = \{e_1, e_2, \dots, e_k\}$}
        	\KwOut{Word-image dictionary $\mathcal{D}$ where each word is associated with a group of images} 
        	\BlankLine
        	\For{each sentence-image pair $(X_i,e_i)_{i \in \{1,2,\dots,k\}} \in (S,E)$}
        	{
        	Filter stop words in the sentence $X_i$\\
        	Segment the sentence by a specified tokenizer that depends on downstream tasks\\
        	The formed subword set is denoted as $T_i=\{t^i_1, t^i_2, \dots, t^i_n\}$\\
        	\For{each word $(t^i_j)_{j \in \{1,2,\dots,n\}} \in T_i$}
        	{
        	Add $e_i$ to the corresponding meaning terms $\mathcal{D}[t^i_j]$ for word $t^i_j$
        	}
        	\Return Word-image dictionary $\mathcal{D}$
        	}
         \end{algorithm*}

\subsection{Multimodal Perception} 
As a special kind of \textit{language} shared by people worldwide, visual modality may help machines have a more comprehensive perception of the real world. Recently, there has been a great deal of interest in integrating image presentations into pre-trained Transformer architectures \citep{su2019vl,lu2019vilbert,tan2019lxmert,li2019unicoder,zhou2019unified,sun2019videobert}. The common strategy is to take a Transformer model \citep{NIPS2017_7181}, such as BERT, as the backbone and learn joint representations of vision and language in a pre-training manner inspired by the mask mechanism in pre-trained language models \citep{devlin2018bert}. All these works require the annotation of task-dependent sentence-image pairs, which are limited to vision-language tasks, such as image captioning and visual question answering. Notably, Glyce \citep{wu2019glyce} proposed incorporating glyph vectors for Chinese character representations. However, it can only be used for Chinese and only involves single image enhancement. \citet{zhang2020neural} proposed using multiple images for neural machine translation (NMT), based on a text-image lookup table trained over a sentence-image pair corpus. However, the number of images is fixed because of the lack of similarity measurement in the simple lookup method, so that it would possibly suffers from the noise of irrelevant images. Compared with \citet{zhang2020neural}, this paper mainly differs by both sides of motivation and technique. 

1) The lookup table constructed at sentence-level as that in \citet{zhang2020neural} would lead to the relatively limited coverage of the related images for each sentence, thus may suffer from noise. Different from incorporating the visual modality as the coarse sentence-level guidance, we propose to leverage word-level multimodal assistance, which offers more accurate means for disambiguation as word-level clues may conveniently help solve sentence-level ambiguity. Meanwhile, multimodality derived from word representations can also be more helpful to provide as many images as possible to alleviate the scarcity of images retrieved for sentences. 

2) Technically, we extend the word embeddings with discriminative image features, enabling the resultant model to benefit broad NLP tasks more than machine translation as previous work focuses on. According to the analysis in Section \ref{analysis}, we see that in most of our datasets, over 50\% tokens have paired images, effectively addressing the drawback of sentence-level method in \cite{zhang2020neural} as it is hard to guarantee adequate images for each sentence. Further, the results in Table \ref{tbl:nmt} show that our method is indeed better than the previous methods, especially in the small-scale datasets, indicating that our method would be useful for low-resource scenarios.

As a result, the major contributions of this work are as follows:

1) Instead of retrieving images for each sentence sparsely, this work focuses on word-level modeling, by enriching word representations with related images as fine-grained visual hints. 

2) The proposed approach does not rely on large-scale aligned sentence-image corpus and can achieve modest results with only a few seed images.

3) Our method is light-weight and is not limited to specific tasks; it is generally motivated to apply visual guidance to a wide range of NLP tasks and can easily be applied to other NLP models.

\section{Model}\label{sec:model}
In this section, we will first introduce the word-image dictionary for word-image(s) retrieval, and then elaborate the proposed visual representation method to enhance word representations with visual guidance.

\subsection{Word-image Dictionary}
During the preprocessing, we set up a dictionary for subsequent word-to-images queries inspired by \citet{zhang2020neural}. Specifically, as described in Algorithm \ref{alg}, the default input setting of the multimodal dataset (i.e., \textit{Multi30K}) is a sentence-image pair denoted as $(X_i,e_i)_{i \in \{1,2,\dots,k\}} \in (S,E)$, where $S=\{X_1, X_2, \dots\, X_k\}$ is the set of input sentences and $E=\{e_1, e_2, \dots, e_k\}$ is the set of paired images. Both sets have the same length $k$. The input sentence $X_i$ is first filtered by a stop word list\footnote{\url{https://github.com/stopwords-iso/stopwords-en}.} and is further segmented to subwords by a specified tokenizer that depends on downstream tasks, which ensures the token overlap with task datasets to the maximum extent. We denote the formed token sequence of $X_i$ as $T_i=\{t^i_1, t^i_2, \dots, t^i_n\}$. Then, for each subword token $t^i_j,j \in \{1,2,\dots,n\}$, we map the paired image $e_i$ to  $t^i_j$ in the dictionary $\mathcal{D}$. At the end of processing the whole multimodal corpus, we form a word-image dictionary where each subword contains diverse images sorted by the number of image occurrence, which are later converted to the image representation as visual hints.\footnote{Examples of paired images for tokens are shown in the Appendix \ref{appendix:exp}.}

\begin{figure}
	\centering
	\includegraphics[width=0.48\textwidth]{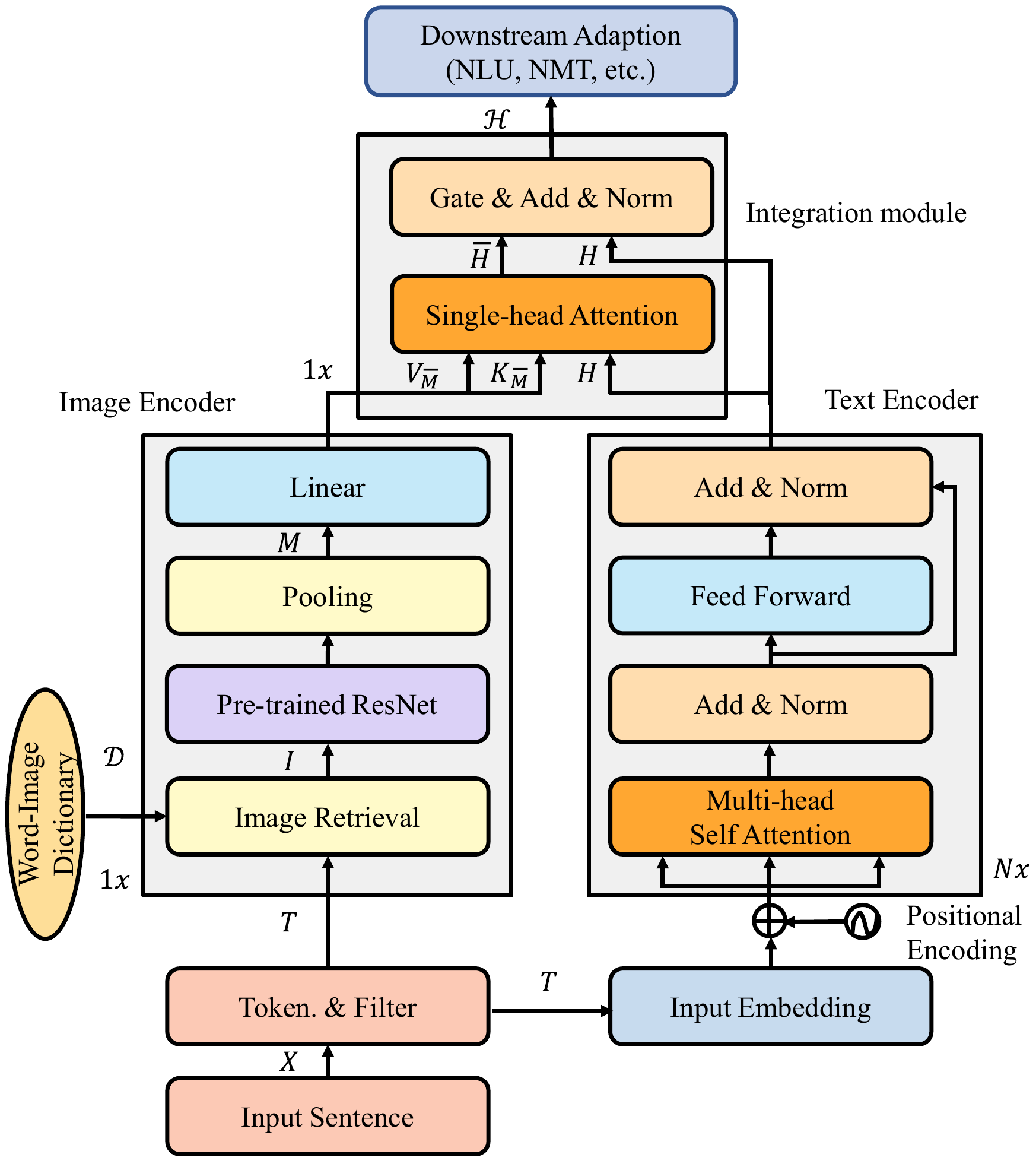}
	\caption{\label{fig:integrate_framework} Overview of our proposed method.}
\end{figure}

\subsection{Integration of the Word and Image Representations}
Figure \ref{fig:integrate_framework} overviews the framework of our visual representation method. First, we employ the same preprocessing method as making the word-image dictionary, that is to filter stop words and tokenize the input sentence $X$ into a sequence of subwords $T=\{t_1, t_2, \dots, t_n\}$. The consistency of preprocessing method guarantees the maximum word overlap between the task datasets and the word-image dictionary. After that, the subword sequence $T$ is fed into a specified text encoder that depends on downstream tasks to learn the source text representation:
\begin{equation}
\begin{split}
   T &= \textup{Tokenizer}(\textup{Filter}(X)),\\ \bm{H}&= \textup{Encoder}(T),
\end{split}
\label{eq1:TextEncoder}
\end{equation}
where $\bm{H}=\{h_1,h_2,\dots,h_n\}$ is the text representation of length $n$.Synchronously, to obtain the corresponding image representation, we look up the word-image dictionary to retrieve top $m$ images for each subword in $T$ as the input to the image encoder. The image encoder is composed of a pre-trained ResNet and a pooling layer:
\begin{equation}
\begin{split}
    I = \mathcal{D}[T], M = \textup{Pooling}(\textup{ResNet}(I)),
\end{split}
\label{eq2:ImageEncoder}
\end{equation}
    where $I=\{I_1, I_2,\dots,I_n\}$ in which $I_i=\{img^i_1,img^i_2,\dots,img^i_m\}$ are the sequence of retrieved images for the $i$-th token, $\mathcal{D}$ is the word-image dictionary and $M=\{M_1, M_2,\dots,M_n\} \in \mathbb{R}^{n \times m \times 2048}$ is the source image representation of dimension $2048$ as fined-grained visual guidance. 
    
    Then, in aim of extracting the region of interest from image representation, on the top of a linear transformation which reduces the dimension of image representation to that of text representation, we apply an attention mechanism following \citet{zhang2020neural} to model the interactions of the text and image representations, which takes text representation as query and image representation as key and value:
\begin{equation}
\begin{split}
    \overline{M}&=\textup{Linear}(M), \\
    \overline{\vh}_i &= \textup{ATT}_{\overline{M}_i}(\vh_i, \bm{K}_{\overline{M}_i}, \bm{V}_{\overline{M}_i}),\\
\end{split}
\label{eq3:Distill}
\end{equation}
where $(\vh_i,{\overline{M}_i})_{i \in \{1,2,\dots,n\}} \in \textit{zip}(\bm{H},\overline{M})$ are respectively the source text, image representation of subword $t_i$, and \{$\bm{K}_{\overline{M}_i}$, $\bm{V}_{\overline{M}_i}$\} are packed from the image representation ${\overline{M}_i}$. Then, we have $\overline{\bm{H}}=\{\overline{\vh}_1,\overline{\vh}_2,\dots,\overline{\vh}_n$\} as the text-conditioned image representation. 
Finally, we apply a gated aggregation method to fuse text representation and image representation in concern of learning the weights of the two independently:
\begin{equation}
\begin{split}
\lambda&=\sigma(\textup{Linear}(\textup{Concat}(\vh_i,\overline{\vh}_i))),\\ \bm{\mathcal{H}}_i&=(1-\lambda)\vh_i+\lambda\overline{\vh}_i,
\end{split}
\label{eq4:Integration}
\end{equation}
where $\textup{Concat}$, $\textup{Linear}$ and $\sigma$ are respectively a concatenation layer, a linear transformer and a logistic sigmoid function. $\lambda \in [0, 1]$ is to weight the expected importance of image representation for the source token. The joint representation $\bm{\mathcal{H}}=\{\bm{\mathcal{H}}_1, \bm{\mathcal{H}}_2, \dots, \bm{\mathcal{H}}_n\}$ will be fed into downstream tasks for prediction.

\subsection{Application in specific NLP tasks}
As a general approach, the visual guidance can be easily applied to standard NLP models. Here, we introduce the specific implementation parts of our proposed method for downstream tasks by taking natural language understanding (NLU) and NMT tasks as examples. For NLU, the baseline model is BERT \citep{devlin2018bert}, and we apply the BERT Tokenizer for subword segmentation. The pooled representation of $\bm{\mathcal{H}}$ will be fed to a feed-forward layer to make the prediction, which follows the same downstream procedure as BERT. For NMT, the text encoder is designated as a self-attention based encoder with multiple layers \citep{NIPS2017_7181}, and the byte pair encoding algorithm is adopted for tokenization. The fused representation $\bm{\mathcal{H}}$ will be directly fed to the decoder to predict the target translation. 

\section{Experiments}
In this section, we first introduce our evaluation tasks and model implementations. For the experiments, we start by presenting experiments on a disambiguation task, and then further conduct a wide range of evaluations on the major natural language understanding and translation tasks, involving 12 NLP benchmark datasets for natural language inference (NLI), semantic similarity, text classification, and machine translation. Part of the NLU tasks are available from the recently released GLUE benchmark \citep{wang2018glue}, which is a collection of nine NLU tasks.

\subsection{Tasks}
\subsubsection{Natural Language Understanding}\label{sec:nlu}
The NLU task involves natural language inference, semantic similarity, and classification subtasks.

\noindent \textbf{Natural Language Inference}
involves reading a pair of sentences and assessing the relationship between their meanings, such as entailment, neutral, or contradiction. We evaluated the proposed method on four diverse datasets: SNLI \citep{Bowman2015A}, MNLI \citep{nangia2017repeval}, QNLI \citep{Rajpurkar2016SQuAD}, and RTE \citep{bentivogli2009fifth}.

\noindent \textbf{Semantic Similarity}
aims to predict whether two sentences are semantically equivalent. Three datasets were used: Microsoft Paraphrase Corpus (MRPC) \citep{dolan2005automatically}, Quora Question Pairs (QQP) dataset \citep{chen2018quora}, and Semantic Textual Similarity benchmark (STS-B) \citep{cer2017semeval}.

\noindent \textbf{Classification}
CoLA \citep{warstadt2018neural} is used to predict whether an English sentence is linguistically acceptable. SST-2 \citep{socher2013recursive} provides a dataset for sentiment classification that needs to determine whether the sentiment of a sentence is positive or negative.

\subsubsection{Neural Machine Translation}\label{sec:nmt}
% We used three translation datasets, WMT'16 English-to-Romanian (EN-RO), WMT'14 English-to-German (EN-DE), and WMT'14 English-to-German (EN-FR), which are standard corpora for NMT evaluation. These datasets cover different sizes of sentence pairs. There are 0.6M, 4.43M, and 36M sentence pairs as the training data, respectively.
The proposed method was evaluated on three widely-used translation tasks, including Multi30K for WMT'16 and WMT'17 multimodal tasks, WMT'14 English-to-German (EN-DE) and WMT'16 English-to-Romanian (EN-RO) for text-only NMT, which are standard corpora for machine translation evaluation.

\noindent \textbf{Multi30K} contains 30K English$\rightarrow$\{German, French\} parallel sentence pairs with visual annotations, which is an extension of Flickr30k \citep{brown2003use}. For WMT'16 and WMT'17 tasks, we have two Test sets, test2016 and test2017, with 1,000 pairs for each. For test2016, we report the results on English-Czech (EN-CS). We also report results on the MSCOCO testset of test2017 that has 461 more challenging out-of-domain instances that contain ambiguous verbs \citep{elliott2017findings}. 

\noindent \textbf{WMT'14 EN-DE} 4.43M bilingual sentence pairs of the WMT14 dataset were used as training data, including Common Crawl, News Commentary, and Europarl v7. 
The \textit{newstest2013} and \textit{newstest2014} datasets were used as the Dev set and Test set, respectively.

\noindent \textbf{WMT'16 EN-RO} we experimented with the officially provided parallel corpus: Europarl v7 and SETIMES2 from WMT'16 with 0.6M sentence pairs. We used \textit{newsdev2016} as the Dev set and \textit{newstest2016} as the Test set.

\subsection{Model Implementation}\label{sec:imp}
Since our task involves text understanding and translation, there are two types of model architectures served as our baselines, the NLU model for language understanding and the NMT model for translation. According to our preliminary experiments, we set the default maximum number of images $m$ for each word as 5. More detailed analysis of $m$ is presented in Section 5. Besides standard NLU and NMT baseline models as described below, we also compare to two baselines: 1) \textit{+ Random}: random baseline where each word is randomly paired with images; 2) \textit{+ Params}: the baseline that has the same number of parameters with our proposed method (\textit{+ VG}) by replacing the image features with the default standard normal distribution as commonly used in the initialization of Embedding layer.\footnote{\url{https://pytorch.org/docs/stable/_modules/torch/nn/modules/sparse.html\#Embedding}} 

\subsubsection{NLU Model}
For the NLU tasks, the baseline was BERT \citep{devlin2018bert}.\footnote{\url{https://github.com/huggingface/transformers}.} We used the public pre-trained weights of BERT and followed the same fine-tuning procedure as BERT. 
% We use the WWM (whole-word-mask) version of the pre-trained weights as default due to its more stable performance.
We use BERT base as default.\footnote{Detailed comparisons of different pre-trained weights are shown in the Appendix \ref{appendix}.} The initial learning rate was set in the range \{2e-5, 3e-5\} with a warm-up rate of 0.1 and L2 weight decay of 0.01. The batch size was selected from \{16, 24, 32\}. The maximum number of epochs was set in the range [2, 5]. Text was tokenized using SentencePiece \citep{sennrich-etal-2016-neural}, with a maximum length of 128.\footnote{\url{https://github.com/google/sentencepiece}.} We report the average Dev set accuracy from 5 random runs.

\begin{table}
% 	\resizebox{\linewidth}{!}
\centering\small
	{
	{
	\begin{tabular}{llll}
		\toprule
		\textbf{Model} &  \textbf{Dev}  & \textbf{Test} & \textbf{Challenge} \\ 
		\midrule
        Baseline & 47.32 & 39.61 & 20.66\\
        Ours & 48.56 (\textbf{+1.24}) & 41.54 (\textbf{+1.93}) & 22.86 (\textbf{+2.20}) \\
		\bottomrule
	\end{tabular}
	\caption{Accuracy (here we use BLEU4 scores) of the multimodal disambiguation experiments on WAT'19 English to Hindi dataset.}\label{tbl:disam}} 
	}

\end{table}

\begin{table*}
	\centering\small
% 	\resizebox{\linewidth}{!}
\setlength{\tabcolsep}{9pt}
	{
		\begin{tabular}{llllllllll}
			\toprule
			\textbf{Method} &  \multicolumn{2}{c}{\textbf{Classification}} &\multicolumn{4}{c}{\textbf{Natural Language Inference}} & \multicolumn{3}{c}{\textbf{Semantic Similarity}} \\
			& CoLA & SST-2 & MNLI & QNLI & RTE& SNLI  & MRPC  & QQP & STS-B  \\ 
			&  (mc) & (acc)	& (acc) & (acc) & (acc)  & (acc) & (F1) & (F1) & (pc)\\
			\midrule
		  %  \multicolumn{10}{l}{\emph{Dev set results}} \\
			BERT  & 57.3  & 92.6  & 84.6 &  90.8 &  66.4 & 90.6 & 88.9 & 87.2 & 89.4 \\
			BERT + Random & 58.5 &  92.4 & 83.9 & 89.9 & 62.3 &  90.5& 89.8 &  87.7 &  86.7\\
			BERT + Params & 57.6 & 92.3 & 83.8 & 89.8 &  61.7 & 90.4 & 89.3 & 87.6 & 86.8 \\
			BERT + VG & \textbf{59.4}++  & \textbf{93.0}+  &  \textbf{84.7} & \textbf{91.1}  & \textbf{69.0}++  & \textbf{90.7} & \textbf{89.8}++ & \textbf{88.2}++ & \textbf{89.6}  \\
% 			BERT$_\text{LARGE}$ & 60.8 & 93.3 & 86.3 & 92.4& 71.1 & 91.3 & 89.5 & 88.0 & 89.5\\
% 			BERT$_\text{LARGE}$ + VG   & 62.8 & 93.6 & 86.6 & 92.5 & 73.6 & 91.5& 90.5 & 88.5 & 90.1\\
% 			BERT$_\text{WWM}$   & 63.6 & 93.6  & 87.2 &93.6 & 77.3  & 92.1 & 90.8 & 88.8 &  90.5\\
% 			BERT$_\text{WWM}$ + VG   & \textbf{64.9} & 93.9 & \textbf{87.4} &  \textbf{93.9} &  \textbf{78.5}  & \textbf{92.2} &  \textbf{90.9} & 88.9  &  \textbf{91.4} \\
% 			\midrule
% 			\multicolumn{10}{l}{\emph{Test set results}} \\
% % 			BERT$_\text{BASE}$  & 52.1 & 93.5  & 84.6 &  - &  66.4  && 88.9 & 71.2  & 87.1  \\
% % 			\hdashline
% % 			BERT$_\text{BASE}$ & 51.4 & 92.1 & 84.4 & 90.3 &   67.1& & 88.3 & 71.3 & 85.1  \\
% 			BERT$_\text{BASE}$ \ \ + VG & 50.7 &  93.1 & 84.3 & 90.5  &  66.7 & 91.0 & 84.9 & 71.0 & 85.8  \\
% 			BERT$_\text{LARGE}$ + VG   & 57.4 & 94.5 & 85.4 &  92.9 &  70.1  &  91.1 & 88.1 & 71.4 & 87.3  \\
%  			% Baseline   & 61.1 &93.6 &86.8 & 93.7 &   77.3 & 91.5 & 87.0 & 71.7 & 88.1 \\
% 			BERT$_\text{WWM}$ \, + VG   & \textbf{61.6} & 94.9 & \textbf{87.1} & \textbf{94.0}  & \textbf{78.6} & \textbf{91.7} & \textbf{90.6} & \textbf{72.6} & \textbf{88.8}  \\
			\bottomrule
		\end{tabular}
	}
	\caption{\label{tab:glue} Results on GLUE benchmark. \emph{mc} and \emph{pc} denote the Matthews correlation and Pearson correlation, respectively. ``++/+" indicate that the proposed method was significantly better than the corresponding baseline at significance level $p$$<$0.01/0.05.
% 	VG denotes our method as \textit{visual guidance}. 
	}
\end{table*}

\begin{table*}
\centering\small
% 	\resizebox{\linewidth}{!}
\setlength{\tabcolsep}{6pt}
	{
\begin{tabular}{llllllllll}
\toprule
\multirow{2}{*}{\textbf{Model}}  & \multicolumn{3}{c}{\textbf{Multi30K 2016}} & \multicolumn{2}{c}{\textbf{Multi30K 2017}} & \multicolumn{2}{c}{\textbf{Multi30K 2017}} & \textbf{WMT'14} &  \textbf{WMT'16}  \\
& & &  & \multicolumn{2}{c}{\textbf{(flickr)}} & \multicolumn{2}{c}{\textbf{(mscoco)}} & \\
& EN-DE & EN-FR & EN-CS & EN-DE & EN-FR & EN-DE & EN-FR  & EN-DE & EN-RO \\
\midrule
\multicolumn{8}{c}{\textit{Existing NMT systems}}                        \\ 
Trans.   & N/A & N/A & N/A  & N/A  & N/A & N/A   & N/A & 27.30 & N/A  \\  
MMT  &  35.09     &   57.40 &  N/A &    27.10   &  N/A &   48.02  &  N/A & N/A &  N/A \\ 
UVR   & 35.72 & 58.32 & N/A & 26.87 & 48.69 & N/A & N/A  & \textbf{28.14} & 33.78 \\
\hline
\multicolumn{7}{c}{\textit{Our systems}}        \\ 
Trans.  & 36.02  &   57.88 & 30.08 & 28.26 &  49.66 & 27.16  & 41.73  & 27.31 & 32.66 \\
Trans. + Random & 36.16 & 58.87 & 29.79 & 28.36 & 50.62 & 27.12 & 41.64 & 27.22 & 32.46\\
Trans. + Rarams & 36.06 & 57.62 & 28.56 & 28.06  & 50.24 & 26.98 & 42.14 & 27.16 & 32.51\\
Trans. + VG & \textbf{36.69}+  &  \textbf{59.47}++ &  \textbf{30.77}+ & \textbf{30.74}++ & \textbf{51.75}++ & \textbf{27.53}+  & \textbf{42.99}++  & 27.80+  & \textbf{33.81}++ \\
\bottomrule
\end{tabular}
}
\caption{BLEU scores on MMT and NMT tasks. Trans. is short for the transformer \citep{NIPS2017_7181}. The MMT and UVR are from \cite{zhang2020neural}. ``++/+" after the BLEU score indicate that the proposed method was significantly better than the corresponding baseline Transformer at significance level $p$$<$0.01/0.05.
}
\label{tbl:nmt}
\end{table*}

\subsubsection{NMT Model}
Our NMT baseline was a text-only Transformer \citep{NIPS2017_7181}. We used six layers for both encoder and decoder. The number of dimensions of all input and output layers was set to 512 (the \textit{base} setting in \citet{NIPS2017_7181}). The inner feed-forward neural network layer was set to 2048. The heads of all multi-head modules were set to eight in both encoder and decoder layers. The learning rate was varied under a warm-up strategy with 8,000 steps. In each training batch, a set of sentence pairs contained approximately 4096$\times$4 source tokens and 4096$\times$4 target tokens. For the Multi30K dataset, we trained the model up to 10,000 steps, and the training was early-stopped if \emph{dev} set BLEU score did not improve for ten epochs. The dropout rate was 0.15. For the EN-DE and EN-RO tasks, the training steps are 200,000. The signtest~\citep{collins-koehn-kucerova:2005:ACL} is a standard statistical-significance test. All experiments were conducted with \textit{fairseq} \citep{ott2019fairseq}.\footnote{\url{https://github.com/pytorch/fairseq}.}

\subsection{Disambiguation Experiment}
A natural intuition of using visual clues for text representation is the advantage of alleviating the ambiguation of language. To evaluate the model performance for disambiguation, we use a dataset from the HVG \citep{parida2019hindi}, which served as a part of WAT'19 Multimodal Translation Task.\footnote{\url{http://lotus.kuee.kyoto-u.ac.jp/WAT/WAT2019/index.html}.} The dataset consists of a total of 31,525 randomly selected images from Visual Genome \citep{krishna2017visual} and a parallel image caption corpus in English-Hindi for selected image segments. The training part consists of 29$K$ English and Hindi short captions of rectangular areas in photos of various scenes, and it is complemented by three test sets: development (Dev), evaluation (Test), and challenge test set (Challenge). The challenge test set was created by searching for (particularly) ambiguous English words based on the embedding similarity and manually selecting those where the image helps to resolve the ambiguity. We did not make any use of the images and use the same settings as the experiments on Multi30K.

We employ the above Transformer model as our baseline and strengthen it with visual guidance (described in Section \ref{sec:model}). As the results shown in Table \ref{tbl:disam}, we observe that our model works effectively on the challenge disambiguation set, indicating that the visual information induced by retrieved images allows disambiguation of translation, which inspires us to apply visual modality as word-level auxiliary information for general language representation.

\subsection{Main Results}\label{sec:result}

Tables~\ref{tab:glue}-\ref{tbl:nmt} show the results for the 12 NLU and NMT tasks, respectively. According to the results, we make the following observations:

1) Table~\ref{tab:glue} shows that our method (+ VG) outperforms the baselines consistently, indicating that it is generally helpful for a wide range of NLU tasks.\footnote{Since the test set of GLUE is not publicly available, we conducted the comparison with our baselines on the Dev set. More detailed results including the Test results are presented in the Appendix \ref{appendix}.} The results verified the effectiveness of modeling visual information for language understanding.

\begin{figure*}
  % \centering

\subfigure{
\begin{minipage}[b]{0.5\linewidth}
\setlength{\abovecaptionskip}{0pt}
%\begin{center}
\pgfplotsset{height=5.5cm,width=8cm,compat=1.15,every axis/.append style={thick},every axis legend/.append style={at={(0.95,0.95)}},legend columns=3 row=1} \begin{tikzpicture} \tikzset{every node}=[font=\small]
\begin{axis} [width=8cm,enlargelimits=0.13,legend pos=north west,xticklabels={1,2,3,4,5,6,7}, axis y line*=left, axis x line*=left, xtick={0,1,2,3,4,5,6}, x tick label style={rotate=0},
  ylabel={BLEU}, ymin=26,ymax=29.5,
  ylabel style={align=left},xlabel={The maximum number of images},font=\small]
+\addplot+ [smooth, mark=*,mark size=1.2pt,mark options={mark color=red}, color=red] coordinates { (0,27.32) (1,27.42) (2,27.66) (3,27.55) (4,28.13) (5,28.05) (6,27.44)};
\addlegendentry{\small Our method}
\addplot+[densely dotted, mark=none, color=orange] coordinates {(0, 26.31)(1, 26.31)(2, 26.31)(3, 26.31)(4, 26.31)(5, 26.31)(6, 26.31)};%\label{plot_2}
\addlegendentry{\small Baseline}
\end{axis}
\end{tikzpicture}
%\end{center}
     \caption{\label{numImg}Influence of the maximum number of images paired for each word on the Multi30K EN-DE Test2017.}
   \end{minipage}
}
\subfigure{
\begin{minipage}[b]{0.5\linewidth}
\setlength{\abovecaptionskip}{0pt}
%\begin{center}
\pgfplotsset{height=5cm,width=8cm,compat=1.14,every axis/.append style={thick}} 

\begin{tikzpicture} \tikzset{every node}=[font=\small] 

\begin{axis} [width=8cm,enlargelimits=0.13, xticklabels={CoLA, SST-2, MNLI, QNLI, RTE, SNLI, MRPC, QQP, STS-B}, axis y line*=left, ybar=5pt, axis x line*=left, xtick={0,1,2,3,4,5,6,7,8}, x tick label style={rotate=0},x tick label style={rotate=45},bar width=5pt,
  ylabel={Word-image coverage (\%)},ymin=10, 
  ylabel style={align=left},xlabel={Datasets},font=\small]
\addplot+[blue1] coordinates {(0,55.47) (1,44.94) (2,32.48) (3,33.41) (4,39.39) (5,63.68) (6,42.67) (7,32.59) (8,49.36)};
\label{plot_one}
% \addlegendentry{BLEU}
\end{axis}
\end{tikzpicture}

%\end{center}
     \caption{\label{overlap} Token overlap (ratios) of the task datasets and the seed Multi30K dataset.}
   \end{minipage}
}

\end{figure*}

2) Results in Table \ref{tbl:nmt} show that our model also outperformed the Transformer baseline in both multimodal (Multi30K) and text-only machine translation (WMT) tasks. As seen, the proposed method significantly outperformed the baseline, demonstrating the effectiveness of modeling visual information for NMT. In particular, the effectiveness was adapted to the translation tasks of the different language pairs, which have different scales of training data, verifying that the proposed approach is a universal method for improving translation performance.

3) Our method introduced a few parameters over our baselines. Taking the NMT model as an example, the extra trainable parameter number is only +4.2M, which is around 6.3\% of the baseline parameters as we used the fixed image embeddings from the pre-trained ResNet feature extractor. Besides, the training time was basically the same as the baseline model. Since both of the \textit{+ Random} and \textit{+ Param} have the same number of parameters with our proposed method \textit{+ VG}, our method still outperform those augmented baselines. The comparison indicates our method does not simply benefit from more parameters. Meanwhile, we notice that \textit{+ Random} showed slightly better performance than the baseline on the Multi30K dataset though it might involve some noise, which is reasonable due to the possible regularization effect and would alleviate over-fitting on the small-scale dataset \citep{brown2003use,noh2017regularizing,brownlee2019train}.

\section{Analysis}\label{analysis}
\paragraph{The influence of numbers of paired images}
To investigate the influence of the maximum number of images $m$ paired for each word, we constrained $m$ in $\{1,2,3,4,5,6,7\}$ for experiments on the EN-DE Test2017 set, as shown in Figure \ref{numImg}. All models outperformed the baseline Transformer (base), indicating the effectiveness of visual guidance. As the number of images increases, the BLEU score generally showed an upward trend at the beginning from 27.32 to 28.13 and dropped sightly after reaching the peak when $m=5$.

\paragraph{Image coverage analysis}\label{sec:coverage}
One possible drawback of the sentence-level method in \cite{zhang2020neural} would suffer from relatively limited coverage of the related images because it is hard to guarantee adequate images for each sentence, which could be alleviated by our fine-grained word-level method. To investigate how many tokens in our tasks can be paired with related images, we calculate the token overlap ratio between the tokens from the multimodal seed dataset and the task datasets. Figure \ref{overlap} depicts the statistics. We see that in most of our datasets, over 50\% tokens have paired images, which indicates that our tasks could enjoy adequate overlap through our method. 

\paragraph{The influence of using different images corpora}
We are interested in whether extra large-scale seed image data could render better model performance. Therefore, we evaluate the performance by further using the larger-scale MS COCO image caption dataset \citep{lin2014microsoft}. The BLEU score of Multi30K EN-DE Test2017 is boosted from 28.13 to 28.36. We contemplate that additional data may further improve the performance, even that image-only data can also be annotated by image caption models and then employed to enhance the model capability, which is left for future work.

\section{Conclusion}
In this paper, we present a visual representation method to explicitly enhance conventional word embedding with multiple-aspect senses by the visual modality. Empirical studies on a range of NLP tasks verified the effectiveness, especially the advance for disambiguation. 
The implementation of our method in the existing deep learning NLP systems demonstrates its flexibility and versatility. In future work, we consider investigating non-parallel text and image data to improve the language representation ability of deep learning models.

\bibliography{acl2020}
\bibliographystyle{acl_natbib}

\appendix
\section{Appendix}

\subsection{Retrieved image examples}\label{appendix:exp}
\begin{figure*}[htb]
	\centering
	\includegraphics[width=1.0\textwidth]{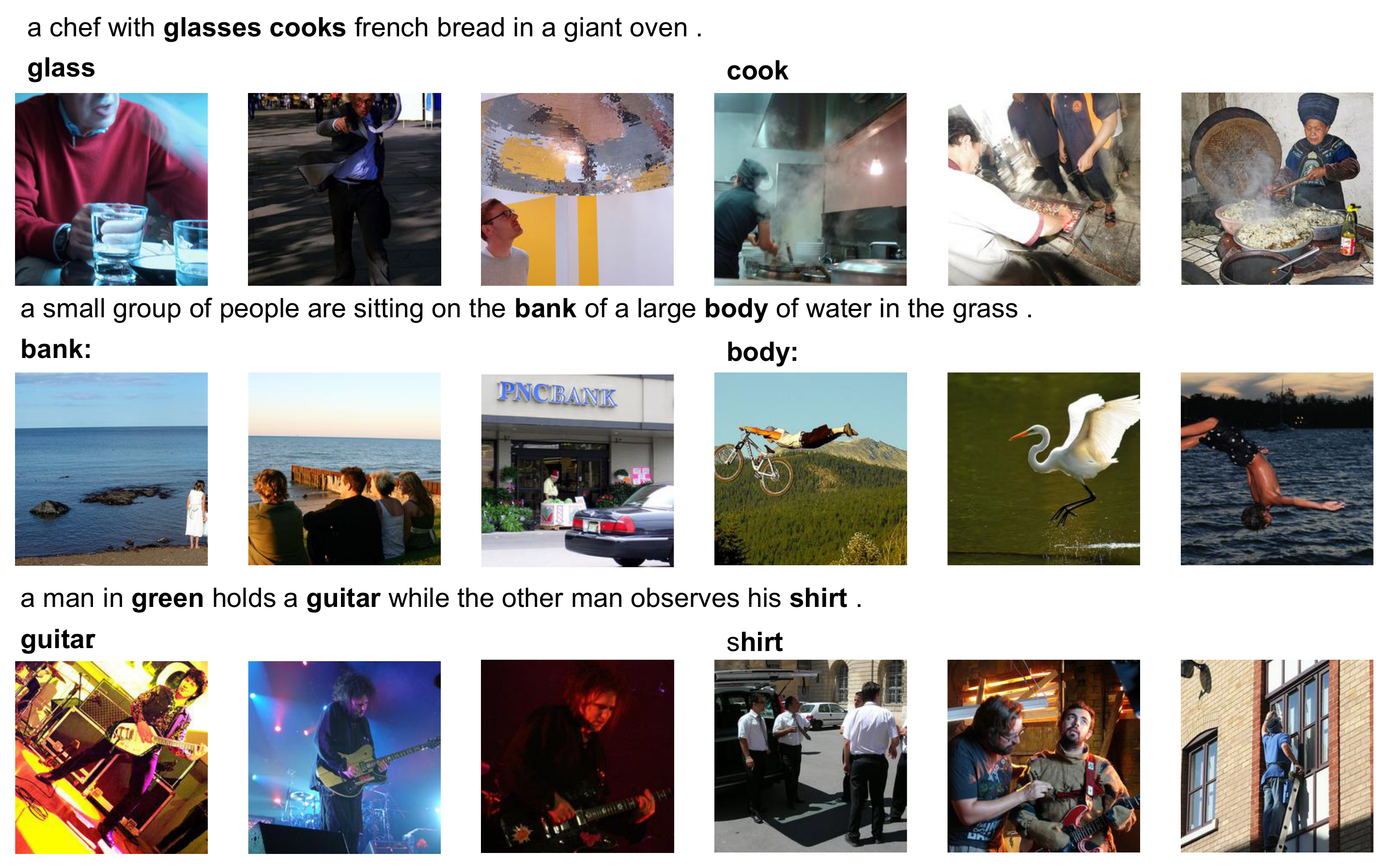}
	\caption{\label{fig:appendix:exp} Retrieved image examples.}
\end{figure*}

Figure \ref{fig:appendix:exp} shows examples to interpret the image retrieval process intuitively, where the words in bold face contain various meanings, for example: \\
\textbf{glass}: a drinking container or an optical instrument;\\
\textbf{cook}: an action or a profession; \\
\textbf{bank}: financial establishment or the land alongside or sloping down to a river or lake; \\
\textbf{body}: physical structure of a person or a body of water; \\
\textbf{guitar}: different shapes and colors; \\
\textbf{shirt}: different shapes and colors. \\

\subsection{Grammatical ambiguity}\label{appendix-amb}
\begin{figure*}[!htb]
\begin{minipage}[t]{0.5\textwidth}
\centering
\includegraphics[width=0.9\linewidth]{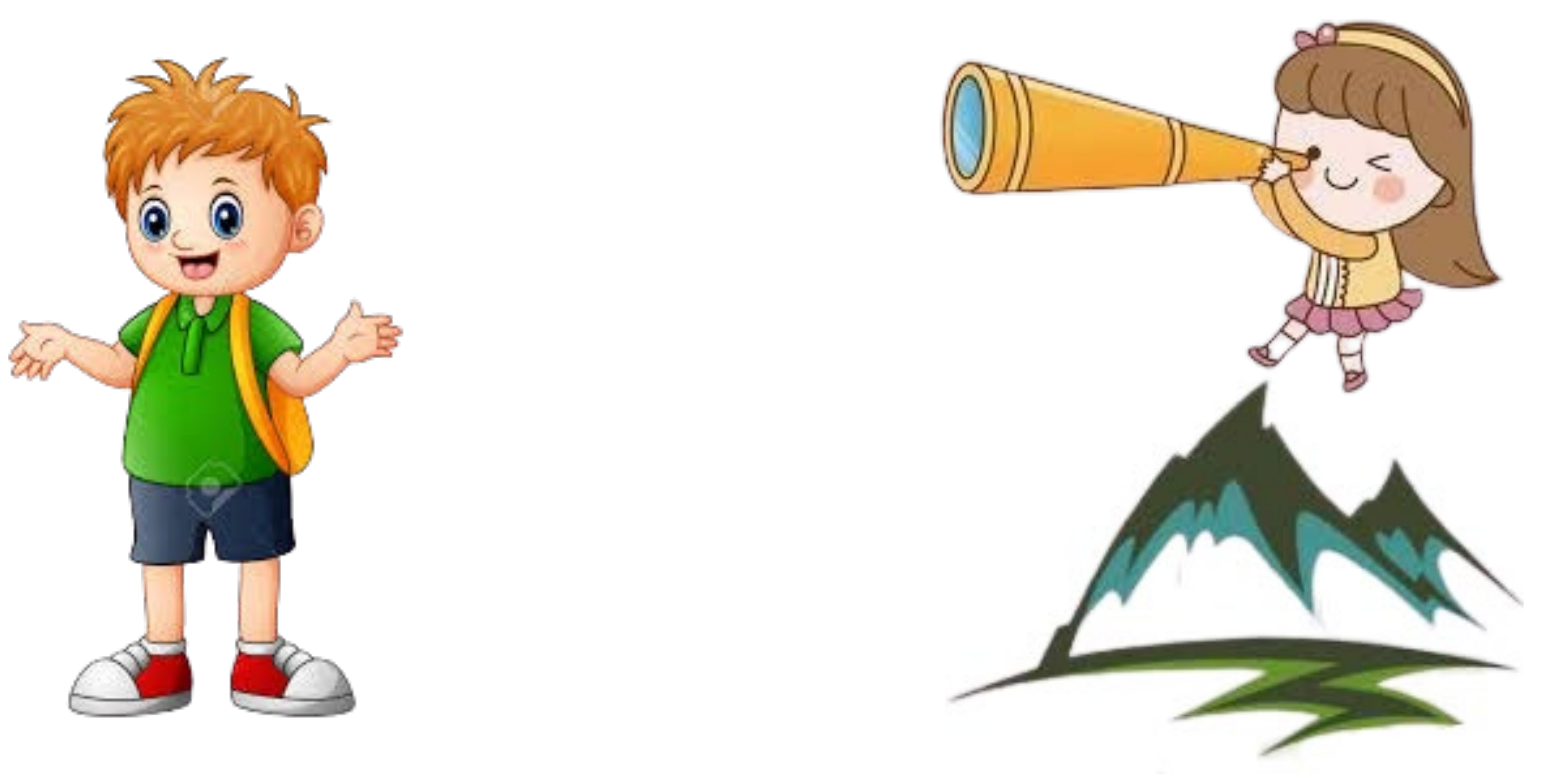}
\caption{\textbf{the girl} saw the man \textbf{on the hill with} \newline \textbf{a telescope}}
\label{fig:shortcoming1}
\end{minipage}
\begin{minipage}[t]{0.5\textwidth}
\centering
\includegraphics[width=0.9\linewidth]{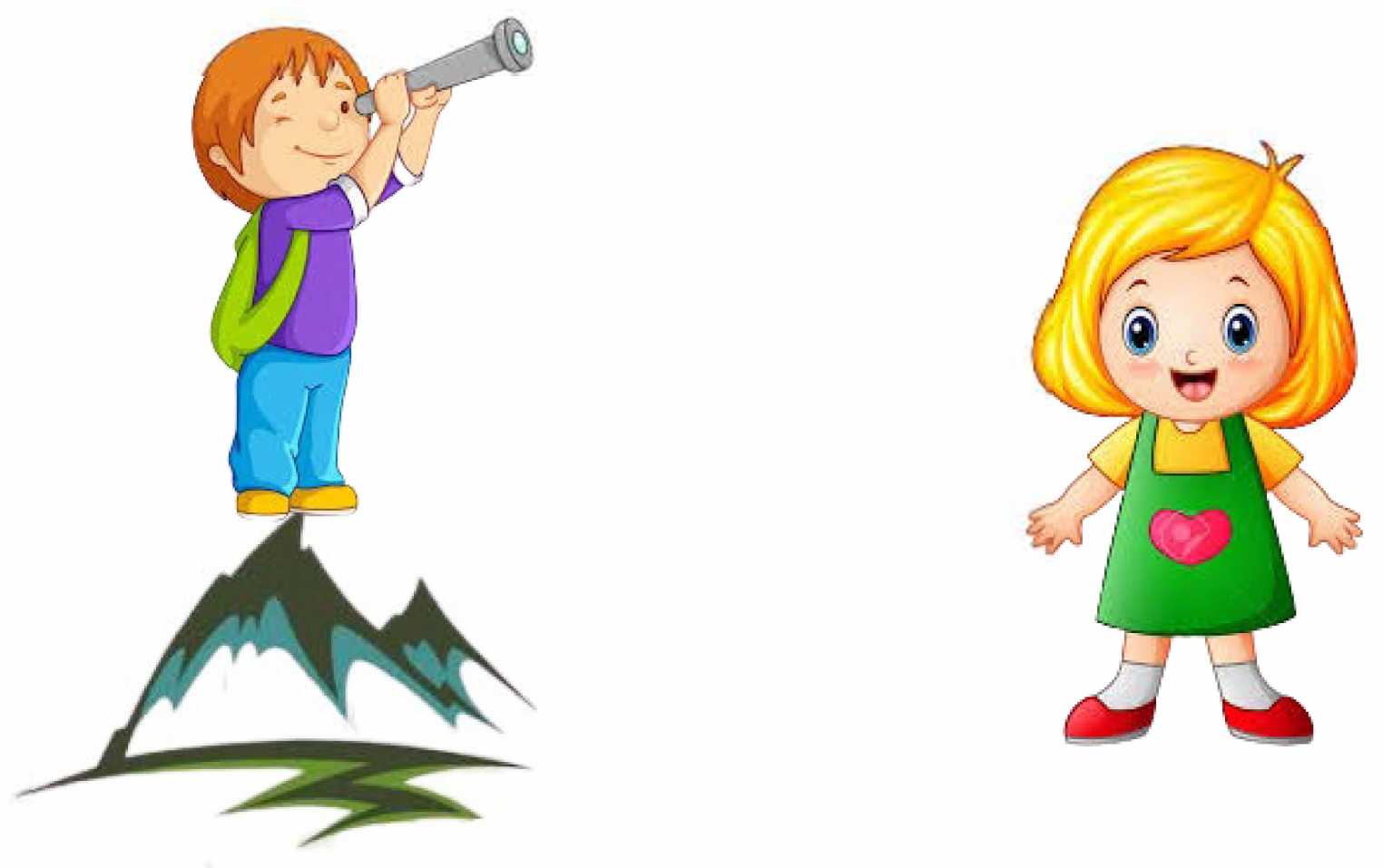}
\caption{the girl saw \textbf{the man on the hill with a telescope}}
\label{fig:shortcoming2}
\end{minipage}
\end{figure*}

Grammatically ambiguous sentences are not covered by our model, such as \textit{the girl saw the man on the hill with a telescope} shown in Figure \ref{fig:shortcoming1} and \ref{fig:shortcoming2}.

\subsection{GLUE results}\label{appendix}
Table \ref{tab:full_glue} shows the complete Dev and Test results for the GLUE benchmark.

\begin{table*}[htb]
	\centering\small
% 	\resizebox{\linewidth}{!}
\setlength{\tabcolsep}{7.2pt}
	{
		\begin{tabular}{lccccccccc}
			\toprule
			\textbf{Method} &  \multicolumn{2}{c}{\textbf{Classification}} &\multicolumn{4}{c}{\textbf{Natural Language Inference}} & \multicolumn{3}{c}{\textbf{Semantic Similarity}} \\
			& CoLA & SST-2 & MNLI & QNLI & RTE& SNLI  & MRPC  & QQP & STS-B  \\ 
			&  (mc) & (acc)	& (acc) & (acc) & (acc)  & (acc) & (F1) & (F1) & (pc)\\
			\midrule
		    \multicolumn{10}{l}{\emph{Dev set results for Comparison}} \\
			BERT$_\text{LARGE}$ & 60.6 &93.2 &86.6 &92.3&	70.4 & 91.0 &88.0 &88.0 &90.0  \\
			MT-DNN &63.5 & \textbf{94.3} &87.1 & 92.9 &	83.4 & 92.2 &87.5  & \textbf{89.2} &90.7 \\
			\hdashline
			
			BERT$_\text{BASE}$  & 57.3  & 92.6  & 84.6 &  90.8 &  66.4 & 90.6 & 88.9 & 87.2 & 89.4 \\
			BERT$_\text{BASE}$ + VG & 59.4  & 93.0  & 84.7 & 91.1  & 69.0  & 90.7 & 89.8 & 88.2 & 89.6  \\
			BERT$_\text{LARGE}$ & 60.8 & 93.3 & 86.3 & 92.4& 71.1 & 91.3 & 89.5 & 88.0 & 89.5\\
			BERT$_\text{LARGE}$ + VG   & 62.8 & 93.6 & 86.6 & 92.5 & 73.6 & 91.5& 90.5 & 88.5 & 90.1\\
			BERT$_\text{WWM}$   & 63.6 & 93.6  & 87.2 &93.6 & 77.3  & 92.1 & 90.8 & 88.8 &  90.5\\
			BERT$_\text{WWM}$ + VG   & \textbf{64.9} & 93.9 & \textbf{87.4} &  \textbf{93.9} &  \textbf{78.5}  & \textbf{92.2} &  \textbf{90.9} & 88.9  &  \textbf{91.4} \\
			\midrule
			\multicolumn{10}{l}{\emph{Test set results for single model with standard single-task training}} \\
			GPT& 45.4 & 91.3  & 82.1 & 88.1 & 56.0 & 89.9 & 82.3  & 70.3 & 82.0  \\
			GPT on STILTs & 47.2 & 93.1   & 80.8 & 87.2  & 69.1& - & 87.7 & 70.1 &85.3    \\
			BERT &60.5 & 94.9 &  86.7 & 92.7 &   70.1 & - & 89.3 & 72.1 & 87.6   \\
			MT-DNN & 61.5 & \textbf{95.6}   & 86.7 & - &  75.5 & 91.6 & \textbf{90.0} & 72.4 & 88.3   \\
			\hdashline
% 			BERT$_\text{BASE}$  & 52.1 & 93.5  & 84.6 &  - &  66.4  && 88.9 & 71.2  & 87.1  \\
% 			\hdashline
% 			BERT$_\text{BASE}$ & 51.4 & 92.1 & 84.4 & 90.3 &   67.1& & 88.3 & 71.3 & 85.1  \\
			BERT$_\text{BASE}$ \ \ + VG & 50.7 &  93.1 & 84.3 & 90.5  &  66.7 & 91.0 & 84.9 & 71.0 & 85.8  \\
			BERT$_\text{LARGE}$ + VG   & 57.4 & 94.5 & 85.4 &  92.9 &  70.1  &  91.1 & 88.1 & 71.4 & 87.3  \\
 			% Baseline   & 61.1 &93.6 &86.8 & 93.7 &   77.3 & 91.5 & 87.0 & 71.7 & 88.1 \\
			BERT$_\text{WWM}$ \, + VG   & \textbf{61.6} & 94.9 & \textbf{87.1} & \textbf{94.0}  & \textbf{78.6} & \textbf{91.7} & \textbf{90.6} & \textbf{72.6} & \textbf{88.8}  \\
			\bottomrule
		\end{tabular}
	}
	\caption{\label{tab:full_glue} Results on GLUE benchmark. The public results are from GPT \citep{radford2018improving}, BERT \citep{devlin2018bert}, MT-DNN \citep{liu2019multi}. \emph{mc} and \emph{pc} denote the Matthews correlation and Pearson correlation, respectively. 
% 	VG denotes our method as \textit{visual guidance}. 
	}
\end{table*}

\end{document}